\documentclass[11pt]{article}

\usepackage[final]{acl}

\usepackage{times}
\usepackage{latexsym}

\usepackage[T1]{fontenc}

\usepackage[utf8]{inputenc}

\usepackage{microtype}

\IfFileExists{inconsolata.sty}{\usepackage{inconsolata}}{}

\usepackage{graphicx}

\usepackage{booktabs}   
\usepackage{amsmath}    
\usepackage{amssymb}    
\usepackage{multirow}
\usepackage{url}

\newcommand{\ead}{\textsc{ead}}   
\newcommand{\mos}{\textsc{mos}}   

\title{Two Emojis of Difference: What Multilingual Affective Generation\\
       Benchmarks Actually Measure}

\author{\bf Fardeen Sadab\thanks{\ \ Equal contribution.}, 
{\bf Adib Sakhawat\footnotemark[1]}
\\Department of Computer Science and Engineering\\
Islamic University of Technology, Dhaka, Bangladesh\\
\texttt{\small\{sakhadib, fardeensadab\}@gmail.com}\\
}

\begin{document}
\maketitle

\begin{abstract}
We audit a multilingual affective generation benchmark---eight instruction-tuned LLMs
producing emoji summaries for 17{,}100 Bangla, English and Hindi sentences, with 6{,}960
human judgements---and find its headline conclusions to be artefacts of the measurement
instrument rather than properties of the systems. Treating annotators as a random rather
than a fixed factor, no system differs significantly from any other
($F(7,14)=0.59$, $p=0.76$), although the conventional analysis declares 19 of 28
pairwise differences significant. Annotator identity explains far more rating variance
than system identity, and the winning system changes whenever any single annotator is
removed. The ordering that does emerge tracks output length: mean emoji count explains
78.7\% of between-system variance, and a within-item length-matched comparison over
2{,}599 pairs reverses the leaderboard. We further show that cross-provider anisotropy
differences vanish under mean-centring, that per-language token costs change sign with
the normalising unit, and that multi-view row-wise splits inflate macro-F1 by $3.1$
points and change the top-ranked system. In place of preference scoring we propose
\emph{emoji--affect decodability}, a reference-based probe whose rankings are stable to
$\pm0.003$ macro-F1 across seeds.
\end{abstract}

\section{Introduction}

Affective computing for low-resource languages is increasingly evaluated by asking
large language models to generate something---a label, a rationale, an emoji---and then
asking a small panel of native speakers which output they prefer. The pattern is
attractive: it needs no gold annotation in the target language, it produces a clean
leaderboard, and it appears to centre the judgements of the speech community rather than
those of a distant annotation vendor.

This paper reports what happened when we subjected one such benchmark to a full
measurement audit. The benchmark is our own: eight instruction-tuned LLMs generate
one to five emojis for each of 17{,}100 sentences in Bangla, English and Hindi, and
three fluent Bangla speakers rate 290 sentences $\times$ 8 systems on a five-point
scale, yielding a completely crossed design with 6{,}960 judgements and no missing
cells. Our initial analysis of this benchmark reported a leaderboard, a
``consistency paradox'' between human preference and emoji--label agreement, provider
differences in embedding anisotropy, a per-language token premium, and near-certain
data leakage under naive splits.

Every one of those findings survives re-examination only in a substantially weakened
form, and several reverse. The failures are not exotic. They are the standard
failure modes of small-panel preference evaluation, applied to a setting---emoji
generation in a low-resource language---where the effect being measured is small, the
annotator population is heterogeneous, and there is no external anchor. We think the
resulting picture is more useful than the leaderboard we set out to produce, because
the same design is now common across low-resource affective NLP.

\paragraph{Contributions.}
\begin{enumerate}\itemsep2pt
\item A generalisability-theory analysis of a fully crossed
$290\times8\times3$ preference study (\S\ref{sec:reliability}). Treating annotators as
a random factor, no system differs significantly from any other; the conventional
analysis finds 19 of 28 pairwise differences significant. We give the
decision study: no number of items yields a reliable ranking with three annotators.
\item An explanation, not merely an observation (\S\ref{sec:length}). Output length
accounts for 78.7\% of the between-system preference variance; a within-item
length-matched comparison over 2{,}599 pairs reverses the ordering. The
``smaller-model-wins'' effect reported previously is a verbosity effect.
\item \emph{Emoji--affect decodability} (\ead{}), a reference-based, annotator-free
metric (\S\ref{sec:ead}), with cross-lingual transfer results showing that the emoji
code models emit is language-specific: transferring a decoder across languages costs
$0.252$ macro-F1 despite $>99.7\%$ vocabulary overlap.
\item Corrections to three widely reported geometric and economic claims
(\S\ref{sec:geometry},\S\ref{sec:cost}): provider anisotropy gaps are a mean-offset
artefact; language separability is not alignment when corpora are not parallel; and
per-language token premia change sign with the choice of normalising unit.
\item An empirical, rather than asserted, quantification of multi-view leakage
(\S\ref{sec:leakage}): $+3.1$ macro-F1 points of optimism and a change in the
top-ranked system.
\end{enumerate}

\section{Related Work}

\paragraph{Human evaluation and its reliability.}
Concerns about the statistical treatment of human judgements in NLP are long-standing
\citep{clark-etal-2021-thats,card-etal-2020-little}, and the specific error we
document---treating annotators as a fixed rather than a random factor---has been raised
for machine translation and summarisation
\citep{graham-etal-2015-accurate,mathur-etal-2020-tangled}. Generalisability theory
\citep{cronbach1972dependability,brennan2001generalizability} provides the standard
apparatus and has been applied only sporadically in NLP. Product-moment correlation is
still frequently reported as an agreement statistic despite being insensitive to
systematic rater bias; \citet{krippendorff2018content} and \citet{shrout1979intraclass}
give the appropriate coefficients.

\paragraph{Length and verbosity bias.}
Human and model-based preference judgements are known to favour longer outputs
\citep{zheng2023judging,dubois2024length,singhal2024long}. Length-controlled variants of
preference metrics have accordingly been proposed for open-ended text. We show the same
bias operating over a radically shorter output space---one to five emojis---where it
accounts for almost all apparent system quality.

\paragraph{Emoji, affect and low-resource languages.}
Emoji carry affective content that is partly conventional and partly culturally
specific \citep{novak2015sentiment,barbieri-etal-2018-semeval,shoeb-de-melo-2020-emotag}.
Bangla affective NLP has grown rapidly \citep{islam-etal-2022-emonoba,das-etal-2021-emotion},
and code-mixed South Asian resources such as EmoMix-3L
\citep{raihan-etal-2024-emomix} highlight transliteration and script noise as failure
sources. Our contribution is not a new emoji resource but a measurement protocol.

\paragraph{Embedding geometry.}
\citet{ethayarajh-2019-contextual} introduced the expected-cosine measure of
anisotropy; \citet{gao2021simcse,li-etal-2020-sentence} link it to representation
quality, and \citet{rudman-etal-2022-isoscore} show that several popular anisotropy
statistics are confounded with the embedding mean. Cross-lingual alignment is
conventionally evaluated by parallel-sentence retrieval
\citep{artetxe-schwenk-2019-massively,feng-etal-2022-language}; we argue that
alignment claims made without parallel data measure something else entirely.

\paragraph{Contamination and leakage.}
Multi-view leakage is a special case of the contamination problem
\citep{sainz-etal-2023-nlp,golchin2024time}. Grouped splitting is standard practice in
clinical and speech corpora but is frequently omitted in multi-model NLP benchmarks.

\section{Benchmark and Study Design}
\label{sec:setup}

\paragraph{Corpora.} We use 17{,}100 sentences: 5{,}813 Bangla, 6{,}547 English and
4{,}740 Hindi, each carrying an emotion label from its source corpus. A design
constraint that we make explicit, and that our initial analysis did not,
is that \emph{the three corpora do not share a label scheme}. Bangla and English use
\{Anger, Fear, Joy, Love, Sadness, Surprise\}; Hindi adds Neutral and Disgust and omits
Love. Bangla is multi-label (17.7\% of items carry more than one label) while English
and Hindi are single-label by construction. Any cross-lingual comparison of label
statistics therefore measures the annotation guidelines of three different corpora, not
the behaviour of a model. All cross-lingual analyses below are restricted to the five
emotions common to all three corpora and to single-label items, giving 4{,}206 Bangla,
5{,}380 English and 3{,}518 Hindi sentences.

\paragraph{Generation.} Eight systems---Claude-3-Haiku, DeepSeek-V3.2,
Gemini-2.0-Flash, Gemma-3-27B, Mistral-Large-2512, GPT-4.1-nano, Qwen3-VL-235B and
Grok-4-Fast---were queried through a single API gateway with an identical
English-language prompt for all languages, \texttt{temperature}$=0.7$,
\texttt{max\_tokens}$=50$, no system prompt, and up to three retries when the response
contained no non-ASCII character. The prompt requests one to five emojis and nothing
else; \emph{the models are never shown the emotion label and are never asked to predict
one}. We restate this because it delimits what the benchmark can support: it is a study
of affective \emph{expression} through a constrained symbolic channel, not of emotion
classification. The full prompt is in Appendix~\ref{app:prompt}.

\paragraph{Human study.} Three independent fluent Bangla speakers rated a stratified 5\% sample of
the Bangla data (290 sentences) for all eight systems on a 1--5 scale of how well the
emojis captured the sentence's emotion. The design is completely crossed: every
annotator rated every (sentence, system) pair, giving $290\times8\times3=6{,}960$
judgements with no missing cells. This is the strongest possible design for the
analyses in \S\ref{sec:reliability}, and it is what allows us to separate the
annotator$\times$system interaction from residual noise at all.

\paragraph{Embeddings.} Each sentence was embedded with five commercial encoders:
Gemini-Embedding-001 (3072d), Mistral-Embed-2312 (1024d), OpenAI
text-embedding-3-large (3072d) and -small (1536d), and Qwen3-Embedding-8B (4096d).
Geometry analyses use a label-stratified sample of 1{,}202 sentences
(400/401 per language); we report the sample size wherever it constrains a conclusion.

\section{Preference Scores Do Not Identify a Winner}
\label{sec:reliability}

\subsection{Agreement}

Pairwise Pearson correlations between annotators are $0.17$, $0.18$ and $0.17$, values
which we initially reported as evidence of ``low but positive
agreement''. Pearson correlation is not an agreement coefficient: it is invariant to
additive and multiplicative rater bias, and here the raters differ enormously in
location, with mean ratings of $2.86$, $4.35$ and $3.31$. Recomputing with coefficients
that penalise systematic bias gives Krippendorff's $\alpha_{\text{ordinal}}=0.014$
(95\% CI $[-0.011,0.039]$, 2{,}000 unit-bootstrap replicates) and
$\text{ICC}(2,1)=0.116$, $\text{ICC}(2,k)=0.282$. For the annotator pair whose Pearson
correlation is $0.166$, $\alpha_{\text{ordinal}}=-0.109$: they are, after accounting for
their different scale usage, in slightly worse than chance agreement. Individual
judgements in this task carry almost no shared signal.

\subsection{The system effect disappears under the correct error term}

\begin{table}[t]
\centering\small
\begin{tabular}{lrrr}
\toprule
Source & MS & $F$ & $p$ \\
\midrule
System $s$ \emph{(vs.\ residual)}   & 11.85 & 18.62 & $<10^{-16}$ \\
System $s$ \emph{(vs.\ $s\times r$)}& 11.85 & \textbf{0.59} & \textbf{0.76} \\
Rater $r$                            & 1345.16 & 319.03 & $<10^{-16}$ \\
System $\times$ rater                & 20.25 & 31.82 & $<10^{-16}$ \\
Residual $p\times s\times r$          & 0.64 & --- & --- \\
\bottomrule
\end{tabular}
\caption{Three-way ANOVA on the crossed $290\times8\times3$ design. The system effect is
significant against the residual and non-significant against the
system$\times$rater interaction, which is the correct error term when annotators are a
sample from a population. The interaction mean square exceeds the system mean square.}
\label{tab:anova}
\end{table}

Table~\ref{tab:anova} gives the three-way ANOVA. The choice of error term decides the
result. Testing the system effect against the residual---which is what one does
implicitly by running a paired test over items on the panel-averaged score, the
near-universal practice in NLP---gives $F(7,4046)=18.62$, $p<10^{-16}$. Testing it
against the system$\times$rater interaction, which is the correct error term if the
three annotators are a sample from a population of possible annotators rather than the
objects of study, gives $F(7,14)=0.59$, $p=0.76$. The interaction mean square
($20.25$) is larger than the system main effect ($11.85$): annotators disagree about
systems more than systems differ.

The same contrast appears in the pairwise comparisons. Over the 28 system pairs, a
paired test over items finds 19 significant at $p<.05$ and 13 after Holm correction;
a test that treats the annotator as the unit of replication finds \emph{zero}
significant at either level. Only 5 of 28 pairs have a difference of consistent sign
across all three annotators.

\subsection{Variance components and the decision study}

Figure~\ref{fig:reliability}(a) decomposes the variance attributable to the random
factors. Annotator identity accounts for 29.1\%, the item$\times$annotator interaction
for 22.9\%, residual noise for 32.6\% and item difficulty for 7.3\%. The system is a
fixed factor and so does not appear in that decomposition; placing it on a common scale
via the sums of squares in Table~\ref{tab:anova}, the annotator main effect accounts for
22.1\% of the total and the system---the quantity the benchmark exists to
measure---for \textbf{0.7\%}. Which sentence is being rated (18.8\% of the total sum of
squares) matters more than twenty times as much as which frontier LLM produced the
emojis.

Because the design is fully crossed we can run a decision study
\citep{brennan2001generalizability}. The generalisability coefficient for the system
mean is
\[
G(n_p,n_r)=\frac{\sigma^2_s}{\sigma^2_s+\frac{\sigma^2_{sr}}{n_r}+\frac{\sigma^2_{ps}}{n_p}+\frac{\sigma^2_{psr}}{n_pn_r}},
\]
where $n_p$ is the number of items and $n_r$ the number of annotators. The current
design attains $G=0.336$. Reaching the conventional threshold $G\ge0.8$ requires 27
annotators at the present 290 items, and 70 annotators for $G\ge0.9$.
Figure~\ref{fig:reliability}(c) makes the important point: because $\sigma^2_{sr}$ is
divided by $n_r$ alone, \emph{the curves saturate in $n_p$}. With three annotators, no
number of sentences---not 5{,}000, not 500{,}000---reaches $G=0.8$. Reliability in this
design cannot be bought with more data; it can only be bought with more annotators.

\subsection{The winner depends on who is on the panel}

\begin{table}[t]
\centering\small
\setlength{\tabcolsep}{4pt}
\begin{tabular}{lccc}
\toprule
Panel & Top system & $\rho$ w/ full & Claude rank \\
\midrule
All three     & Gemma-3-27B      & ---   & 3 \\
$-$A1         & Claude-3-Haiku   & $0.55$ & 1 \\
$-$A2         & Gemma-3-27B      & $0.90$ & 4 \\
$-$A3         & Mistral-Large    & $0.62$ & 8 \\
\bottomrule
\end{tabular}
\caption{Leave-one-annotator-out leaderboards. Removing any single annotator from a
three-person panel can change the winner. Claude-3-Haiku ranges from first to last.}
\label{tab:loo}
\end{table}

\begin{figure*}[t]
\centering
\includegraphics[width=\textwidth]{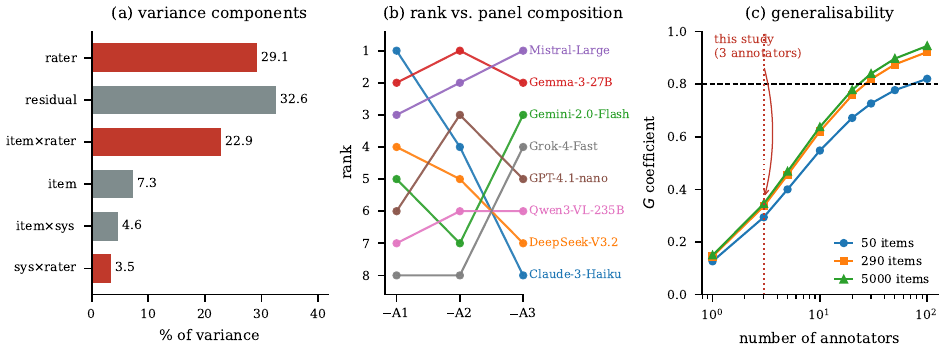}
\caption{(a) Random-effect variance components; annotator-related terms (red) dominate.
The system is a fixed factor and accounts for 0.7\% of the total sum of squares against
22.1\% for the annotator main effect. (b) System rank as a function of which annotator
is excluded.
(c) Decision study: $G$ saturates in the number of items, so three annotators cannot
reach $G=0.8$ at any corpus size.}
\label{fig:reliability}
\end{figure*}

Table~\ref{tab:loo} and Figure~\ref{fig:reliability}(b) show the practical consequence.
Dropping one annotator changes the top-ranked system in two of three cases.
Claude-3-Haiku is ranked first by one panel and last by another. Between-annotator
Kendall correlations over the eight system means are $-0.36$, $0.07$ and $0.00$: at the
level of the leaderboard, the three annotators are not measuring the same thing.

\section{What the Ratings Actually Track}
\label{sec:length}

The obvious next question is \emph{why} the smaller open models outperformed the
larger proprietary ones. The answer is that they emit more emojis.

\begin{table}[t]
\centering\small
\setlength{\tabcolsep}{3.4pt}
\begin{tabular}{lrrrr}
\toprule
System & \mos{} & emojis & \ead{} & win$_{\text{LM}}$ \\
\midrule
Gemma-3-27B      & 3.667 & 4.95 & 0.653 & 0.543 \\
Mistral-Large    & 3.638 & 4.87 & 0.639 & 0.506 \\
Claude-3-Haiku   & 3.579 & 4.92 & 0.584 & 0.499 \\
GPT-4.1-nano     & 3.516 & 4.55 & 0.562 & 0.428 \\
DeepSeek-V3.2    & 3.454 & 3.45 & 0.560 & 0.541 \\
Gemini-2.0-Flash & 3.443 & 2.67 & 0.657 & \textbf{0.603} \\
Qwen3-VL-235B    & 3.421 & 3.88 & 0.597 & 0.422 \\
Grok-4-Fast      & 3.325 & 2.58 & 0.655 & 0.517 \\
\midrule
$\rho$ with \mos{} & --- & $0.90^{**}$ & $-0.19$ & $0.12$ \\
\bottomrule
\end{tabular}
\caption{Systems ordered by raw \mos{}. Mean emoji count tracks \mos{} almost perfectly;
decodability does not. win$_{\text{LM}}$ is the within-item length-matched win rate,
which reorders the table. $^{**}p<.01$.}
\label{tab:length}
\end{table}

\begin{figure*}[t]
\centering
\includegraphics[width=\textwidth]{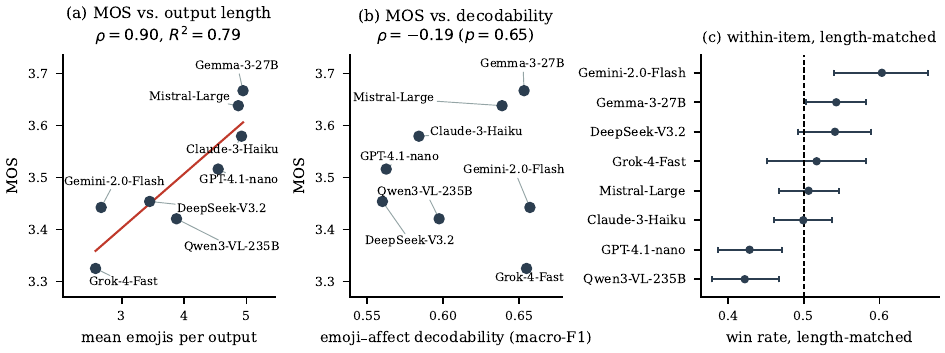}
\caption{(a) Mean output length explains 78.7\% of between-system \mos{} variance.
(b) Emoji--affect decodability does not correlate with \mos{}. (c) Within-item
length-matched win rates reorder the leaderboard; error bars are 95\% binomial CIs.}
\label{fig:length}
\end{figure*}

Table~\ref{tab:length} and Figure~\ref{fig:length}(a) show that the mean number of
emojis a system emits predicts its \mos{} with $\rho=0.90$ ($p=0.002$), accounting for
78.7\% of the between-system variance. A judgement-level mixed model with item and
annotator random effects estimates $\beta=0.159$ \mos{} points per additional emoji
($\text{SE}=0.015$, $p=1.6\times10^{-25}$). The entire spread between the best and worst
of eight frontier systems is $0.341$ \mos{} points---slightly more than two emojis'
worth of output length.

\paragraph{Length-matched comparison.} Adjusting for length statistically is not
sufficient, because the number of emojis a system emits is a system-level property. We
therefore run a comparison that conditions on nothing at the system level: for every
sentence and every pair of systems that happened to emit \emph{the same number of
emojis on that sentence}, we record which was rated higher. This yields 2{,}599 matched
comparisons. The resulting ordering (Table~\ref{tab:length}, Figure~\ref{fig:length}(c))
correlates with the raw \mos{} ordering at $\rho=0.12$ ($p=0.78$). Gemini-2.0-Flash
moves from sixth to first (win rate $0.603$, 95\% CI $[0.540,0.664]$); Grok-4-Fast
moves from last to fourth. The reported ``efficiency--performance'' effect, in which
smaller models were preferred, does not survive.

\paragraph{Who is length-biased?} The per-annotator slope on emoji count is $0.085$
($p<10^{-5}$), $-0.024$ ($p=0.14$) and $0.329$ ($p<10^{-48}$). One annotator's
system-level ranking correlates with output length at $\rho=0.95$; another's at
$\rho=-0.26$. The length bias, like everything else in this study, is an annotator
property that the panel average silently aggregates.

\section{Emoji--Affect Decodability}
\label{sec:ead}

If preference scores cannot rank systems, something else must. We propose a
reference-based metric that uses the emotion labels the source corpora already carry,
requires no annotators, and is defined without reference to any emoji--affect lexicon,
thereby avoiding the unvalidated emoji-to-label mapping our initial analysis relied on.

\paragraph{Definition.} Let $g$ be a generator and $\mathcal{D}_L=\{(x_i,y_i)\}$ the
single-label items of language $L$ over the shared five-emotion set. Let $E_g(x)$ be the
multiset of emoji graphemes $g$ emits for $x$ (ZWJ sequences and skin-tone modifiers
kept intact). \ead{}$(g,L)$ is the 5-fold cross-validated macro-F1 of an
$\ell_2$-regularised multinomial logistic regression over sublinear TF-IDF features of
$E_g(x)$, predicting $y$. It measures how much of the corpus's affective distinction
survives compression into $g$'s emoji channel. A generator that emits the same three
emojis for everything scores at the floor regardless of how pleasant those emojis are.

\paragraph{Properties.} \ead{} is stable: across ten random seeds the largest
per-system standard deviation is $0.0033$ macro-F1, and 2{,}000-replicate item
bootstraps give CIs of width $\approx0.032$ that separate the systems into three
distinct tiers (Figure~\ref{fig:geometry}(c)). It is cheap: no annotators, and
deterministic given a fixed set of model outputs. And it disagrees with human preference:
$\rho=-0.19$ ($p=0.65$) with \mos{}. Per annotator the correlation is $+0.07$, $+0.76$
and $-0.40$. Given \S\ref{sec:reliability}, we do not read this as evidence against
\ead{}; a metric cannot be validated against a criterion that does not reliably
order the systems.

\paragraph{Results.} Bangla \ead{} ranges from $0.560$ (DeepSeek) to $0.657$ (Gemini),
against a majority-class floor of $0.087$; English from $0.515$ to $0.593$; Hindi from
$0.376$ to $0.458$ (Figure~\ref{fig:geometry}(b)). The union of all eight systems'
emojis reaches $0.686$ in Bangla, above any single system, indicating that the
generators encode partly complementary information.

\paragraph{The emoji code is language-specific.} Training the decoder on one language's
emoji outputs and testing on another costs $0.252$ macro-F1 on average, even though
target-language vocabulary coverage exceeds $99.7\%$ in every direction: the models emit
the \emph{same} emojis across languages but use them to mean different things. Emoji
are frequently described as a language-neutral affective interface; under a controlled
test, at least as produced by these systems, they are not.

\paragraph{Matched comparison with sentence embeddings.} On identical items
($n\approx400$ per language), the best emoji channel reaches $0.673$ macro-F1 in Bangla
against $0.595$ for the best sentence embedding, $0.613$ against $0.736$ in English, and
$0.430$ against $0.387$ in Hindi. It is tempting to conclude that for Bangla a
five-emoji summary carries as much affective information as a 4096-dimensional
commercial encoder. We report a control that tempers this: the embedding learning curve
is still rising steeply at $n=400$ ($0.428\to0.480\to0.594$ for $n=100,200,400$) whereas
the emoji curve is nearly flat ($0.614$ at $n=400$, $0.657$ at $n=4{,}206$). The
embedding numbers are therefore lower bounds and we claim only that the emoji channel is
\emph{competitive} at matched sample size, not that it is superior.

\section{Geometry: Two Corrections}
\label{sec:geometry}

\begin{figure*}[t]
\centering
\includegraphics[width=\textwidth]{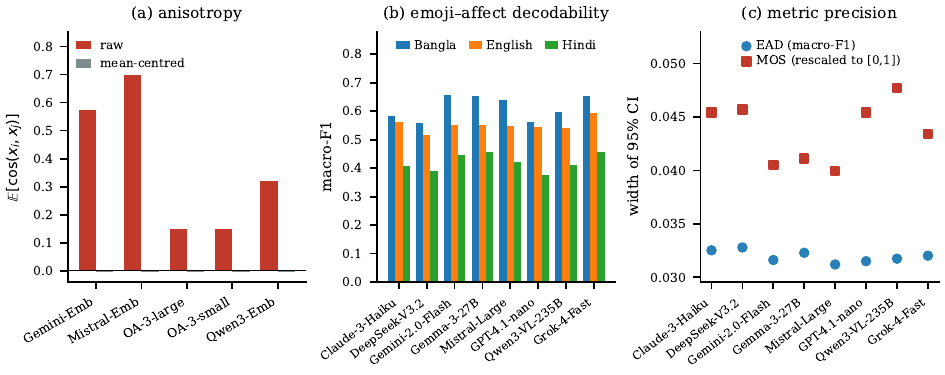}
\caption{(a) Provider anisotropy differences vanish under mean-centring.
(b) Emoji--affect decodability by system and language. (c) Width of the 95\% CI for
\ead{} and for \mos{} rescaled to a common range; \ead{} is the more precise
instrument.}
\label{fig:geometry}
\end{figure*}

\paragraph{Anisotropy differences are a mean offset.} Following
\citet{ethayarajh-2019-contextual} we compute
$A(X)=\mathbb{E}_{i\neq j}[\cos(x_i,x_j)]$, evaluated exactly as
$(\|\sum_i u_i\|^2-n)/(n(n-1))$ for unit-normalised $u_i$. Raw values differ sharply
across providers: $0.149$ (OpenAI-3-small), $0.150$ (OpenAI-3-large), $0.319$ (Qwen3),
$0.574$ (Gemini) and $0.700$ (Mistral)---a range we initially interpreted as a
meaningful quality difference. After subtracting the global mean, every model lies in
$[-0.0008,0.0013]$ (Figure~\ref{fig:geometry}(a)). The differences are entirely
attributable to a common displacement vector, removable by one centring operation, as
\citet{rudman-etal-2022-isoscore} predict. Moreover, raw anisotropy does not predict
downstream utility: its correlation with the pooled emotion-probe macro-F1 across the
five encoders is $r=-0.48$ ($p=0.41$, $n=5$). We can find no sense in which the
anisotropy gap matters.

\paragraph{Language separability is not cross-lingual alignment.} Our initial analysis
reported a ``cross-lingual divergence'' of $0.09$ for one provider against
$0.72$ for another and read the lower value as better alignment. That inference is not
available here, because the three corpora are not parallel: no Bangla sentence is a
translation of any English sentence. Any statistic comparing language-conditional
distributions is therefore confounded with topic, register and domain. What can be
measured is \emph{separability}. A linear language classifier reaches $\ge0.993$
accuracy on four of five encoders and $0.597$ on Gemini-Embedding-001, whose Fisher
separation ratio is correspondingly low ($0.009$ vs.\ $0.105$--$0.280$). But low
separability is not a virtue: Gemini also has the weakest emotion probe of the five
($0.290$ pooled macro-F1 vs.\ $0.593$ for Qwen3). Its space does not so much align
languages as fail to represent the distinctions we tested. Establishing alignment would
require parallel data and a retrieval evaluation
\citep{artetxe-schwenk-2019-massively}, neither of which this corpus supports.

\section{Cost Accounting Changes Sign with the Unit}
\label{sec:cost}

Raw token counts are not comparable across providers with different tokenizers, and the
choice of normalising unit determines the direction of any per-language conclusion.
Measured in prompt tokens per source \emph{byte}, Bangla is the \emph{cheapest} of the
three languages ($1.108$ vs.\ $1.450$ for English and $1.557$ for Hindi). Measured in
prompt tokens per source \emph{character}, Bangla costs $2.93$ against English's
$1.45$---a $2.0\times$ premium---because Bangla text averages $2.64$ UTF-8 bytes per
character. Both statements are true; the ``token premium'' reported previously is a
statement about Unicode encoding, not about tokenizer fairness, and should be reported
per character with the byte ratio disclosed.

A second confound: raw completion-token counts conflate emitted output with internal
reasoning traces. Grok-4-Fast averages $294.7$ completion tokens per Bangla item while
emitting $2.60$ emojis ($124$ tokens per emoji); the other seven systems average
$1.4$--$4.4$. Comparing systems on billed completion tokens compares reasoning
configurations, not task efficiency.

Finally, no cost measure predicts quality. Spearman correlations with \ead{} are
$-0.14$ (total tokens), $-0.12$ (completion tokens) and $-0.14$ (tokens per emoji), none
significant. The only cost-like quantity that predicts \mos{} is the number of emojis
emitted ($\rho=0.90$), which \S\ref{sec:length} shows to be a bias rather than a
quality signal.

\section{Multi-View Leakage, Measured}
\label{sec:leakage}

A benchmark that stores $V$ views of each of $N$ source items (here $V=8$ generators, or
$V=5$ encoders) and is then split row-wise places the same source sentence on both sides
of the partition. The probability that no item is split is $(\rho^V+(1-\rho)^V)^N$,
which for $V\ge2$ and $N\ge100$ is numerically zero at any usual $\rho$; reporting it as
a percentage close to 100, as we initially did, is uninformative. The
decision-relevant quantity is the expected fraction of test rows whose source item also
appears in training, $1-(1-\rho)^{V-1}$: $99.998\%$ at $V=8,\rho=0.8$ and $99.84\%$ at
$V=5$.

We measure the resulting optimism. On the Bangla multi-view table (4{,}206 items
$\times$ 8 views $=33{,}648$ rows), a naive row-wise 80/20 split gives
$0.630\pm0.003$ macro-F1 against $0.598\pm0.014$ for a split grouped by source item:
$+3.1$ points absolute, $+5.3\%$ relative ($t=4.97$, $p=0.0006$, 6 replicates). The
inflation is stable at $1.7$--$2.3$ points across split ratios from $0.5$ to $0.9$. More
importantly it perturbs the leaderboard: system rankings under leaky and clean protocols
correlate at $\rho=0.76$ and the top-ranked system changes from Mistral-Large to
Gemma-3-27B. Grouped splitting is not a hygiene footnote; it changes which system a
benchmark declares best.

\section{Recommendations}

\begin{enumerate}\itemsep2pt
\item \textbf{Report agreement with a bias-sensitive coefficient.} Pearson correlation
between annotators is not agreement. Report Krippendorff's $\alpha$ or an ICC with a
bootstrap CI, and report annotator mean and SD separately.
\item \textbf{Treat annotators as a random factor.} If the claim is about systems in
general rather than about three specific people, the error term must include the
system$\times$annotator interaction. Reporting a paired test over items on
panel-averaged scores is not sufficient.
\item \textbf{Run a decision study before scaling up collection.} A small fully crossed
pilot is enough to estimate $\sigma^2_{sr}$ and hence the panel size the intended claim
requires. If more items cannot fix the design, that is worth knowing before 6{,}960
judgements are collected.
\item \textbf{Length-match, do not length-adjust.} Report a within-item comparison
restricted to outputs of equal length alongside the raw score.
\item \textbf{Report a reference-based metric next to preference.} \ead{} or an analogue
gives a reproducible number that does not move when the panel changes.
\item \textbf{Centre embeddings before reporting anisotropy}; do not describe
non-parallel corpora as evidence of cross-lingual alignment; normalise token counts by
characters and disclose the bytes-per-character ratio; split multi-view tables by source
item.
\end{enumerate}

\section*{Limitations}

The human study has three annotators, all of them university-educated Bangla speakers
from a similar age band; the variance components we report are estimates from
$n_r=3$ and are correspondingly imprecise, and the population of annotators they
generalise to is narrow. A larger and demographically broader panel could plausibly
show smaller annotator variance, though our decision study indicates the panel would
need to be an order of magnitude larger before the current design became reliable. We
report per-annotator results throughout rather than only the panel average, since the
panel average is precisely the quantity our analysis shows to be unstable.

The human study covers Bangla only, so the reliability findings are demonstrated rather
than shown to be universal; we expect them to be worse, not better, for languages with
fewer available annotators. Our three languages are all South Asian or English, and we
avoid claims about ``multilingual affective AI'' in general.

\ead{} inherits whatever biases the source corpora's emotion labels carry, and the three
corpora differ in label scheme, granularity and provenance; we therefore compare systems
only within a language, never across. The metric also rewards systems whose emoji usage
is \emph{consistent} with the corpus, which is not the same as being good, useful or
culturally appropriate---a system that consistently uses a culturally inappropriate
emoji for an emotion would score well. \ead{} replaces the lexicon-based consistency
metric it supersedes, not human judgement properly powered.

The geometry analyses use 1{,}202 sentences and five commercial encoders accessed
through an API; we do not have access to their training data, and API models can change
without notice, so the geometric measurements are properties of the endpoints as we
queried them. The embedding-versus-emoji comparison is sample-limited as described in
\S\ref{sec:ead}.

Finally, this is an audit of a single benchmark that we built ourselves. We believe the
failure modes generalise because the design is conventional, but we have not
demonstrated that they do.

\section*{Ethics Statement}

The three annotators are fluent Bangla speakers who worked independently of the research
team, participated voluntarily, and were informed of the purpose of the study before
beginning. 
The sentences derive from publicly
available Bangla, English and Hindi corpora and were not filtered for personally
identifying content beyond what the source corpora applied, and we work from sentence
identifiers and model outputs under the source corpora's licences. Emotion
attribution to text written by identifiable individuals carries a risk of
misrepresentation, and we caution against deploying emoji-based affect systems in
consequential settings, particularly given that our best system recovers only 66\% of a
five-way emotion distinction in Bangla. Model outputs were not filtered for offensive
emoji use; we observed no such cases but did not audit exhaustively.

%

\bibliography{custom}

\appendix

\section{Prompt}
\label{app:prompt}

Identical for all systems and all three languages, sent as a single user turn with no
system prompt:

\begin{quote}\small\ttfamily
You MUST respond with 1-5 emojis representing the emotion in this sentence.\\[2pt]
CRITICAL: Your response MUST contain emojis. Empty responses are NOT allowed.
Do NOT include any text, words, or explanations - ONLY emojis.\\[2pt]
Sentence: \{sentence\}\\[2pt]
Emojis (REQUIRED):
\end{quote}

Decoding: \texttt{temperature}$=0.7$, \texttt{max\_tokens}$=50$, provider defaults for
\texttt{top\_p} and stop sequences, no seed (the provider gateway does not expose one).
A response containing no character above U+007F triggered a retry, up to three
attempts; observed retry rates were $0.001$--$0.064$ per system.

\section{Analysis Settings}
\label{app:repro}

Every analysis in this paper runs on CPU from the stored model outputs; no further API
access is needed once generation and embedding are complete. Random seeds are fixed
throughout. Bootstrap replicate counts are 2{,}000 for agreement coefficients and
\ead{} rank distributions and 5{,}000 for \mos{} confidence intervals. Probes are
$\ell_2$-regularised multinomial logistic regressions ($C=4$ on sublinear TF-IDF emoji
features, $C=1$ on $\ell_2$-normalised embeddings) under 5-fold stratified
cross-validation, with class-balanced weights throughout.

Emoji are segmented into graphemes, keeping zero-width-joiner sequences and skin-tone
modifiers attached to their base character and discarding the occasional stray ASCII a
model emits despite the prompt. This matters for \S\ref{sec:length}: naive per-codepoint
segmentation inflates the mean emoji count by 15.0\% overall, but unevenly across
systems (from $+5.4\%$ for Claude-3-Haiku to $+31.5\%$ for DeepSeek-V3.2), so it
distorts the between-system length ordering rather than merely rescaling it.

\end{document}